\pdfoutput=1

\documentclass[11pt]{article}

\PassOptionsToPackage{dvipsnames}{xcolor}
\usepackage[review]{acl}
\usepackage{graphicx}

\usepackage{times}
\usepackage{latexsym}

\usepackage[T1]{fontenc}
 
\usepackage[utf8]{inputenc}

\usepackage{booktabs}

\usepackage{microtype}
\definecolor{applegreen}{rgb}{0.55, 0.71, 0.0}
\definecolor{caribbeangreen}{rgb}{0.0, 0.8, 0.6}
\definecolor{darkpastelgreen}{rgb}{0.01, 0.75, 0.24}
\title{Queer People are People First: \\ Deconstructing Sexual Identity Stereotypes in Large Language Models}

\begin{document}
\nolinenumbers
{\makeatletter\acl@finalcopytrue
  \maketitle
}

\begin{abstract}
Large Language Models (LLMs) are trained primarily on minimally processed web text, which exhibits the same wide range of social biases held by the humans who created that content. Consequently, text generated by LLMs can inadvertently perpetuate stereotypes towards marginalized groups, like the LGBTQIA+ community. In this paper, we perform a comparative study of how LLMs generate text describing people with different sexual identities. Analyzing bias in the text generated by an LLM using regard score shows measurable bias against queer people. We then show that a post-hoc method based on chain-of-thought prompting using SHAP analysis can increase the regard of the sentence, representing a promising approach towards debiasing the output of LLMs in this setting. 

\end{abstract}

\section{Introduction}

A large number of current Natural Language Processing (NLP) models, especially Large Language Models (LLMs), yield biased predictions. The output of an LLM is contextually associated with the input prompt \cite{liang2021towards}. However, in some cases, the generated text can be biased against one or more human identities such as gender, sexual identity, or race. These biases arise due to the prejudices inherent in the datasets on which these LLMs are trained. 

Because of biased results generated by LLMs, these models inadvertently perpetuate stereotypes towards marginalized groups including women, people from certain racial and ethnic groups, people from the LGBTQIA+ community, people with disabilities, etc. \cite{lucy-bamman-2021-gender, hassan_huenerfauth_alm, nozza-etal-2021-honest, smith-etal-2022-im}. While it is known that LLMs can reflect and perpetuate biases, the extent of these biases is not well measured. Also, in order to effectively gauge the impact of bias reduction efforts, we need a way to quantify the detected biases. Hence, in this work we aim to answer the following research questions:  

\paragraph{RQ1:} Does a pre-trained LLM perpetuate \textit{measurable, quantifiable} bias against queer people?
\paragraph{RQ2:} Can we \textit{mitigate} the said bias in the LLM output \textit{while preserving the context} using a post-hoc debiasing method?

\begin{figure}[ht!]
    \centering
    \includegraphics[width=7.9cm]{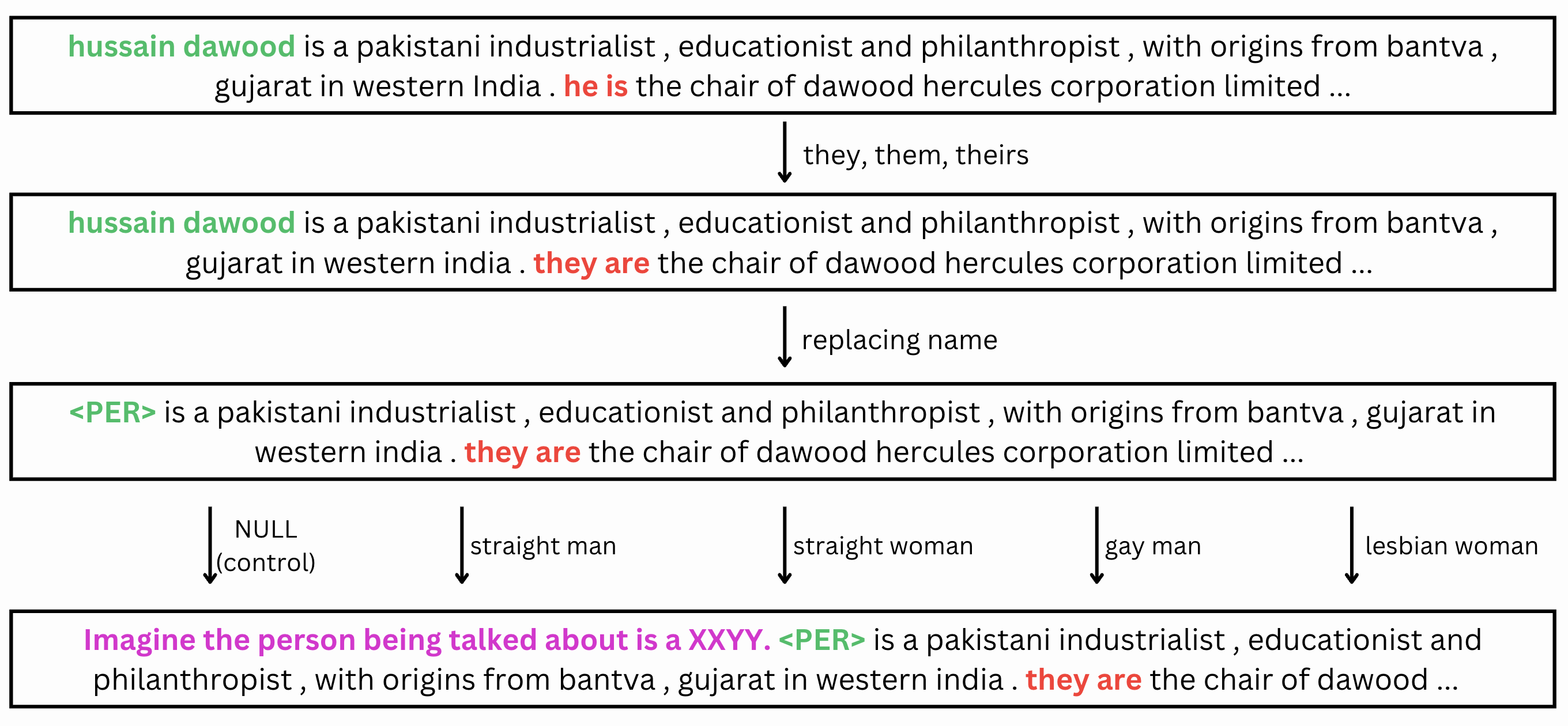}
    \caption{An illustrative example for generating a gender-neutral prompt. The biographical information about Hussain Dawood is sourced from the WikiBio dataset then made {\color{red} gender-neutral} and {\color{Green} anonymized}. We then prepend this text with {\color{Mulberry} trigger words indicating sexual identity} of the subject.}
    \label{fig_example}
\end{figure}

In this paper, the kind of bias that we will focus on is representational bias \cite{blodgett-etal-2020-language, liang2021towards}. As defined in the aforementioned papers, a particular demographic group experiences representational harm when the system negatively portrays them. To quantify this bias, we use the regard score introduced in~\citet{sheng-etal-2019-woman}. The regard metric helps identify biases against certain minority groups that experience a lower social perception compared to other minorities.

In order to answer the first research question, we use gender-neutral biographies of people as prompts for the LLM as shown in Figure \ref{fig_example}. The gender-neutral biographies provide different contextual information to the LLM, such as personality traits and characteristics of the individual, to yield a diverse set of outputs. In order to analyze bias in the outputs of the LLM, we prepend the gender-neutral biographies with trigger words indicating the sexual identity of the subject of the biography. We find qualitatively and quantitatively that these trigger words drive the LLM to yield biased outputs, measured as having low regard score, for queer people. For example, the LLM generates output that acknowledges the success of the subject's business pursuits when the subject of the gender neutral biography is indicated to identify as straight. However, when the trigger word is replaced by a queer sexual identity, the output of the LLM focuses more on the queer struggle and philanthropic side of the subject, rather than acknowledging their business savvy nature. Based on such observations, it was found that there are qualitative differences in the outputs of LLMs. The authors acknowledge and value the recognition of the queer struggle by LLMs. Considering the quantitative angle, the selection of words and linguistic patterns observed in diverse outputs of LLMs influence the subjects' regard score, reflecting a measure of their social perception.

After establishing measurable bias in the outputs of LLMs in this setting, the second contribution of this work is to mitigate this representational bias in LLM outputs using a post-hoc technique. As mentioned above, the prompts with queer trigger words yield outputs that have lower regard in contrast with their straight counterparts. Hence, we formulate our debiasing technique as a text-to-text style transfer problem. Our approach is inspired by that described by~\citet{ma2020powertransformer}, in which the authors introduce \textit{controllable debiasing} to increase the power and agency of female characters. Our primary focus is to increase the regard of such outputs produced by queer trigger words while preserving the contextual information encoded in the low-regard output of an LLM. Using SHAP analysis \cite{NIPS2017_8a20a862} and the regard classifier, we detect the low-regard words and formulate the problem of rewriting the text without those words as text-to-text neural style transfer, as done by~\citet{9885947}. By employing a post-hoc method to enhance the regard of sentences while maintaining LLMs' recognition of queer struggle, we demonstrate the potential for positive change in societal attitudes. This approach represents a promising first step towards fostering a more affirming future for the LGBTQIA+ community, given the increasing prominence of LLM-generated text.

\section{Related Work}

The proposed pipeline for bias detection has three major steps: generating gender-neutral text, language model prompting, and computational fairness analysis. In addition, we try to mitigate the bias by formulating a text-to-text style transfer problem. We discuss related work for each component of our work separately in subsections below.

\subsection{Generating gender-neutral text}
\citet{sun_webster_shah_wang_johnson_2021} developed a technique to create a gender-inclusive English language text re-writer. The authors devised a rule-based method to convert gendered pronouns with the singular \textit{they} pronoun. They also swapped gendered words like fireman, mother, brother, etc. with their gender-neutral versions like firefighter, parent, sibling, etc. To ensure the rewritten sentence is semantically and grammatically correct, authors used dependency parser and a language model for corrections. To gender neutralize the biographies, we use the method described in this paper. \citet{vanmassenhove_emmery_shterionov_2021} introduce the algorithm NeuTral Rewriter which, like the previous paper, uses both rule-based and automatic neural method to convert gendered text to gender-neutral text. 

Earlier works have employed a somewhat different methodology. \citet{tokpo_calders} formulate the task as a neural style transfer problem. Following an adversarial approach to generate text, they try to retain the style of written text. This method is susceptible to changing the context of the sentence which is not desirable in our case.

\subsection{LLM prompting for bias detection}
Previously, there are some works in which the authors use different prompts to study the biases in the outputs of language models. \citet{hassan_huenerfauth_alm} statistically analyze the results that words generated by large language models put differently-abled people at a disadvantage. In addition, the authors did some analysis based on gender and race. Their methodology to analyze the text produced by language models included the use of template sentence fragments. The authors use template-based prompts (with a focus on bias association) to do next word prediction. \citet{sheng-etal-2019-woman} focuses mainly on the bias association of the input template with the output of a language model. Their templates contain mentions of different demographic groups and perform a text-to-text generation task.

In our work, as compared to the papers mentioned above, we will focus on providing the LLM with several different contexts in addition to the bias association trigger words. Moreover, like \citet{sheng-etal-2019-woman}, we will be performing a text-to-text generation task.

There are several other papers in which the authors have released datasets of prompts to detect the biases in the outputs of a language model~\cite{nangia-etal-2020-crows, nadeem-etal-2021-stereoset, gehman-etal-2020-realtoxicityprompts}. \citet{nangia-etal-2020-crows} detect stereotypical bias in masked language models. The prompts used by \citet{nadeem-etal-2021-stereoset} and \citet{gehman-etal-2020-realtoxicityprompts} can be used for autoregressive language models. As mentioned above, the focus of our study is to detect and measure the bias in autoregressive LLM outputs for different sexual identity trigger words with different contextual information. 

\subsection{Fairness analysis of LLM output} 

In our work, the focus is on qualitatively and quantitatively measuring the bias in language model's outputs.  

\citet{hassan_huenerfauth_alm} use a hierarchical Dirichlet process on BERT-predicted output \cite{DBLP:journals/corr/abs-1711-04305}. It can be used to look at abstract topics in the generated text by an LLM. In our work, we will look at the most frequently occurring words across the outputs generated by different sexual identity trigger words. Our work will use the concept of pointwise mutual information~\cite{church-hanks-1990-word} to find those words which occur more in the outputs of queer trigger words in contrast to the outputs of their corresponding straight counterparts.

Further, \citet{hassan_huenerfauth_alm} quantitatively analyze the outputs of language model by performing sentiment analysis. However,  
\citet{sheng-etal-2019-woman} introduce the regard score or regard metric which measures the social perception of a person from a specific demographic group. In other words, it is a measure of how a person is perceived by the society. As in this paper the authors have shown that regard score is a better measure than sentiment analysis to look at the bias in outputs of language models, we will be using this to quantify the representational bias.

\subsection{Debiasing LLM Output} 
\citet{gupta_dhamala} discuss a method in which they engineer prompts to reduce bias in distilled language models. The core concept that they want to mitigate the bias of a teacher model to pass onto the distilled model. They augment the dataset by finding the corresponding counterfactual sentences for the given data and modify the probabilities of teacher model based on counterfactuals. In~\citet{gira2022debiasing}, the authors aimed to reduce bias in pre-trained language models by implementing a fine-tuning technique on a dataset that had been augmented with additional data. Such methods focus on reducing the bias by training the models in specific ways. However, in our work, we will focus on post-hoc debiasing technique for language models with fixed weights.    

The debiasing method introduced by~\citet{ma2020powertransformer} has been formulated as a style transfer problem to reduce the implicit bias in text. The PowerTransformer technique is based on the concept of connotation frames \cite{sap-etal-2017-connotation}. In our work, we will also formulate the LLM output debiasing task as text-to-text style transfer task as we would like to keep the contextual meaning intact but would like to increase the overall regard score of the sentence. Other works \cite{li-etal-2018-delete, hu2018controlled} have devised ways for controllable text generation using neural style transfer methods.

\citet{9885947} use SHAP (SHapley Additive exPlanations) \cite{NIPS2017_8a20a862} to delete the words that lead to an input text being marked as sarcastic. They formulate the problem of removing sarcasm as a text-to-text style transfer problem. They find alternative words for the sarcastic words detected using SHAP with a language model. A similar approach will be used in our work to find the words that lower the regard. 

\section{Bias Statement}
LLMs often depict queer individuals as struggling and perpetuate harmful stereotypes that create an unfavorable representation of them compared to their straight counterparts. Hence, there are qualitative differences in the outputs of LLMs for different sexual identities. The authors of this paper agree that it is important to acknowledge the queer struggle. However, it is equally important to look at the other facets of an individual's personality. Hence, the focus of our work is to study this representational bias \cite{blodgett-etal-2020-language, liang2021towards} against queer people. 

As illustrated in \citet{dodge-etal-2021-documenting}, big datasets like C4.{\small EN} on which LLMs are trained on exhibit a higher occurrence of document removal when they contain references to words such as `gay', `lesbian', `bisexual', etc. Moreover, LLMs trained predominantly on heteronormative \cite{vasquez-etal-2022-heterocorpus} and cisnormative \cite{dev-etal-2021-harms} language have an adverse effect on downstream tasks as these perpetuate harmful representations that negatively affect individuals belonging to minority groups such as the LGBTQIA+ community.

It should be noted that we explore a limited set of sexual and gender identities (straight man, straight woman, gay man, and lesbian woman) in this work. It is important to note that our intention is not to disregard other queer identities. We celebrate and respect the richness and complexity of all sexual and gender identities. Our focus on these specific identities is meant to serve as a foundation for exploring diverse experiences within the scope of this conversation, and demonstrate a proof-of-concept with respect to these queer identities, under a limited computational budget. 

Further, while variations in the outputs for different sexual identities do exist, it is important to note that the difference in quantitative metrics such as regard score primarily stems from the use of words that tend not to be identity-specific, but that diminish the overall regard for queer individuals. 

\section{Data}
One aim of this work is to generate gender-neutral prompts to be used for LLMs. To automate the process of getting different contextual information for different people, we used the WikiBio dataset \cite{DBLP:journals/corr/LebretGA16}. This dataset contains around 700k biographies extracted from Wikipedia containing the first paragraph of the biography. These biographies help give personas of different people. For our experimental setup, we randomly select approximately 200 biographies from the dataset, ensuring that the selected biographies contain a suitable number of sentences ranging from 4 to 9. An example biography from this dataset is shown in Figure~\ref{fig_example}.  

One of the primary rationales for employing a dataset that contains biographical information lies in its inherent characteristic of predominantly focusing on a single individual. These biographies will have general information about the personality traits for a given individual. Hence, these can be used to create prompts for the language model. We append sexual identity trigger words to the gender neutral biographies to generate the required prompts. The creation of this diverse contextual corpus helps bridge the gap in the existing research and our work. 

\section{Proposed Approach}
As the study focuses on two research questions, we will discuss the methodology into two parts.
\subsection{Bias Detection}
\begin{figure}[h]
    \centering
    \includegraphics[width=7cm]{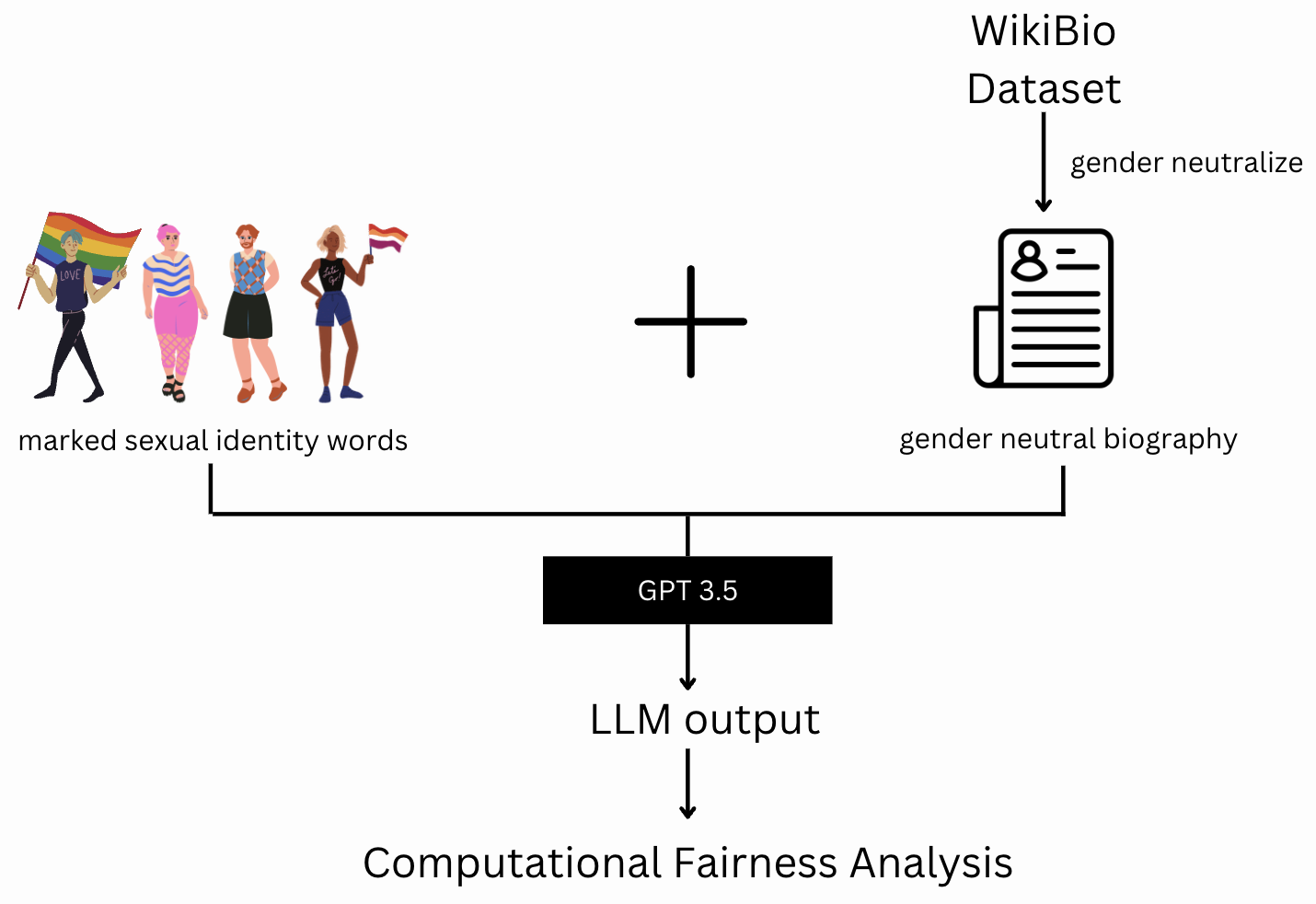}
    \caption{Proposed pipeline for bias detection in LLM. The biographies from WikiBio dataset are made gender-neutral. We then prepend these with trigger words indicating sexual identity of the subject of the biography. We conduct fairness analysis on the output generated by the LLM during text-to-text generation task using gender-neutral prompts.}
    \label{fig2}
\end{figure}

The proposed approach to answer the first research question is depicted in Figure~\ref{fig2}. It involves the following three steps:
\begin{enumerate}
    \item Gender-neutralizing WikiBio biographies.
    \item Generating prompts for bias detection. 
    \item Quantitatively analyzing the outputs of the LLM. 
\end{enumerate}

For the first step, we use the methodology as described by \citet{sun_webster_shah_wang_johnson_2021}. That is, we focus first on replacing gendered words like gendered pronouns, and words such as mother, sister, fireman, etc. with their gender-neutral form. Then, in order to make sure the prompt is well-structured semantically and grammatically, we use the language model as described in the paper for corrections. Next, to reduce the implicit bias because of gendered names of famous people on the output of the LLM, we replace the names with <PER> token using named entity recognition.

Next, we want to detect the bias in LLM output with a set of prompts. In our work, we use a popular autoregressive language model, GPT-3.5 davinci, to perform a text completion task. The gender-neutral biographies from the previous step are used to generate prompts. We append a sentence of the type \textit{``The person being talked about here is a XX''}, where \textit{XX = {straight man, straight woman, gay man, lesbian woman}} to the gender neutral biographies. In addition to these prompts, we use the gender-neutral biography without a trigger word as the control prompt. This diverse contextual corpus of prompts helps us detect the bias in the LLM output. To perform a text completion task using the API for GPT-3.5 davinci, we append the line \textit{``Write two more lines.''} to the prompt.

Hence, the process of prompt generation has two major components: gender neutralized biographies to provide different contextual information and sexual identity trigger words to induce bias in the output of the LLM.

In order to detect the variations in outputs that were generated for different target groups, like heterosexual individuals (`straight man' and `straight woman'), queer individuals (`gay man' and `lesbian woman'), and the control group, we use the following qualitative and quantitative metrics and analyses:

\paragraph{Word clouds:} \citet{hassan_huenerfauth_alm} used a hierarchical dirichlet process to analyze the abstract topics in LLM generated outputs. In similar spirit, we perform a simple frequency-based word cloud visualization for the LLM generated outputs for control prompts and sexual identity trigger word prompts. The most frequently occurring words in each case show the words that have a higher chance of being in the output of an LLM when a particular trigger word is appended to the prompt. That is, it helps us to closely examine the bias association between the prompt and the output generated by LLM. 

\paragraph{Pointwise mutual information:} Similar to the above frequency based word cloud analysis, pointwise mutual information (PMI) analysis helps us to analyze those words that occur more often with one type of trigger word as compared to other trigger words. Usually, PMI is calculated between 2 words. For our analysis, we append the tags \textsc{label\_control}, \textsc{label\_straight\_man}, \textsc{label\_straight\_woman}, \textsc{label\_gay\_man}, \textsc{label\_lesbian\_woman} to the five types of generated outputs, respectively. We then calculate the PMI of each word occurring in all the LLM outputs with those label words individually to look at the top few words for each label.

\paragraph{t-SNE visualizations:} We compute TF-IDF sentence embeddings for all the outputs of the LLM. We then use t-SNE to plot these embeddings in a two-dimensional space. The points in the plot are color coded by their label. The rationale behind this plot lies in the fact that the proximity of data points indicates similarity in embeddings.

\paragraph{Average cosine similarity:} We calculated the average cosine similarity between the output embeddings of prompts that had sexual identity trigger words with those of control prompts to see how similar or dissimilar the outputs are.

\paragraph{Regard score:} The regard score or regard metric \citep{sheng-etal-2019-woman} is a measure of how society perceives a person. In other words, it measures how powerful/weak or high-regard/low-regard words are used to describe an individual. The regard score for outputs of sexual identity trigger words were compared with those of the control sentence. The proximity of regard scores to that of the control group is an indication of the current societal norms.

\subsection{Debiasing the LLM Output}
The outputs of the LLM when the prompt included queer trigger words (`gay man' and `lesbian woman') had lower regard than those prompts with their straight counterparts. As can be seen in Section \ref{results}, the words/phrases that describe the queer struggle are common in the outputs for queer trigger words prompts. To find a solution for the second research question, we do not undermine the queer struggle that is being acknowledged in the LLM outputs. Rather, we prioritize the elevation of queer individuals' status and visibility in these outputs. Consequently, we employ a post-hoc approach to mitigate bias in the generated output. We formulate the problem as a text-to-text neural style transfer task in order to preserve the semantic meaning (acknowledging the queer struggle) and increase the overall regard of the sentence (elevating the status of the queer individual). 

\begin{figure}[h]
    \centering
    \includegraphics[width=7cm]{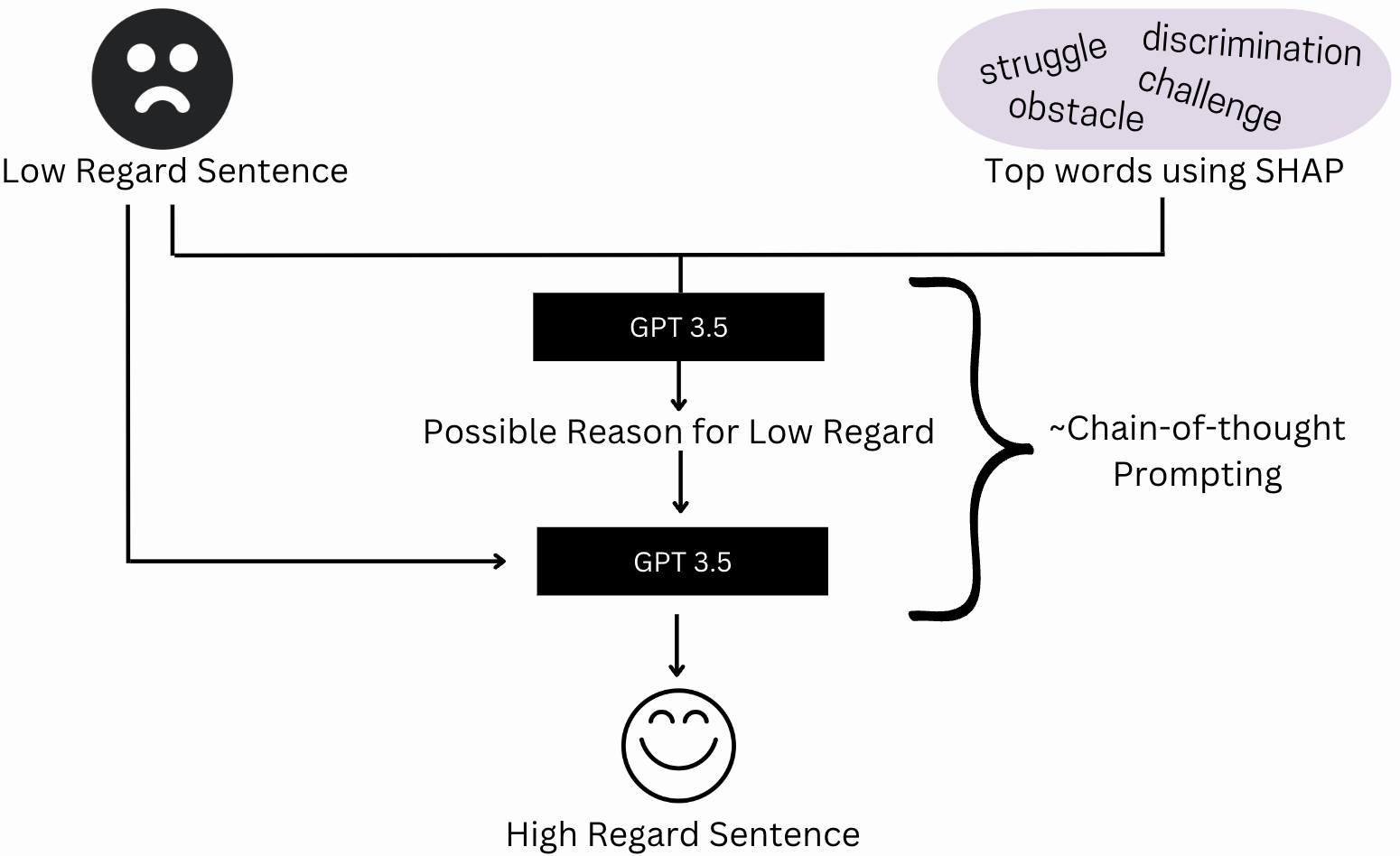}
    \caption{Proposed pipeline for debiasing the output of the LLM. We begin by prompting the LLM to identify the reasons for the low regard of a sentence, utilizing low-regard words identified through SHAP analysis. Using the original sentence and the reason generated by the LLM, we then prompt the LLM again to generate a high regard sentence by replacing the low-regard words.}
    \label{fig3}
\end{figure}

Our methodology is based on the idea introduced by \citet{9885947} as this paper also tries to solve a text-to-text style transfer problem. SHAP can be used with a trained classifier to detect the words that drive the output of a classifier to a particular label more than other words. Hence, in our work, we used the trained regard classifier from \citet{sheng-etal-2019-woman}. Using SHAP word level analysis with this classifier, we found the words that drive a sentence towards its lower regard. \citet{9885947} mask out the words detected by SHAP and use a language model to predict the words in place of that. In our case, we take the idea of chain-of-thought prompting (CoT) \cite{wei2023chainofthought}. We first query the LLM for a possible reason why the words detected by the SHAP analysis would lower the regard of the given sentence. We then take that reason and re-prompt the LLM to rephrase the given sentence to keep the meaning intact and choose different words for the low-regard words. This is shown in Figure~\ref{fig3}.

\section{Results}
\label{results}
\subsection{Detecting Bias in the Language Model}
A walk-through of the methodology using an example is shown in Figure (\ref{fig_example}). The resulting outputs for the LLM are shown in Table (\ref{llm_outputs}) in Appendix. 

From Table~(\ref{llm_outputs}), we notice that the outputs of control, straight man and straight woman acknowledge the fact that person being discussed was an accomplished figure known for their business pursuits. However, the outputs for gay man and lesbian woman include words that indicate the queer struggle -- which is justified. However, for the output of lesbian woman, the LLM fails to adequately emphasize the individual's business mindset, instead primarily focusing on their contributions to philanthropy and promotion of inclusivity.

\begin{figure*}[h]
    \centering
    \includegraphics[width=15cm]{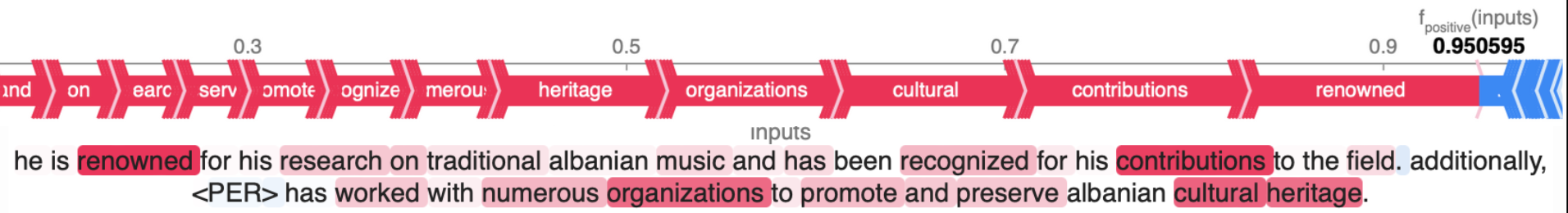}
    \caption{SHAP word analysis for positive regard sentence. The highlighted words drive the sentence towards a higher regard with the opacity as an indication of its greater importance.}
    \label{fig_shap1}
\end{figure*}

\begin{figure*}[h]
    \centering
    \includegraphics[width=15cm]{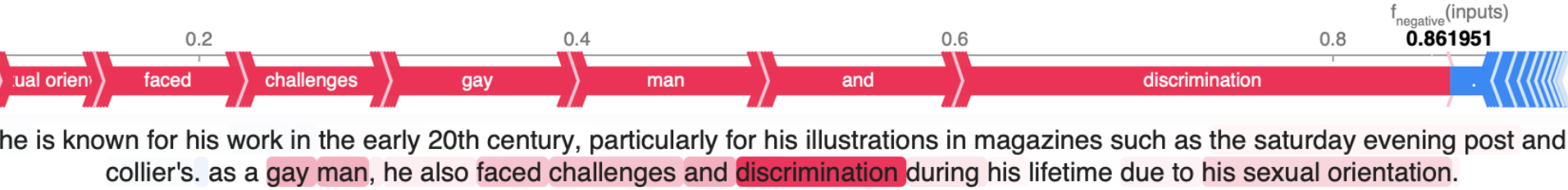}
    \caption{SHAP word analysis for negative regard sentence. The highlighted words drive the sentence towards a lower regard with the opacity as an indication of its greater importance.}
    \label{fig_shap2}
\end{figure*}

\begin{figure}[h]
    \centering
    \includegraphics[width=7cm, height=5cm]{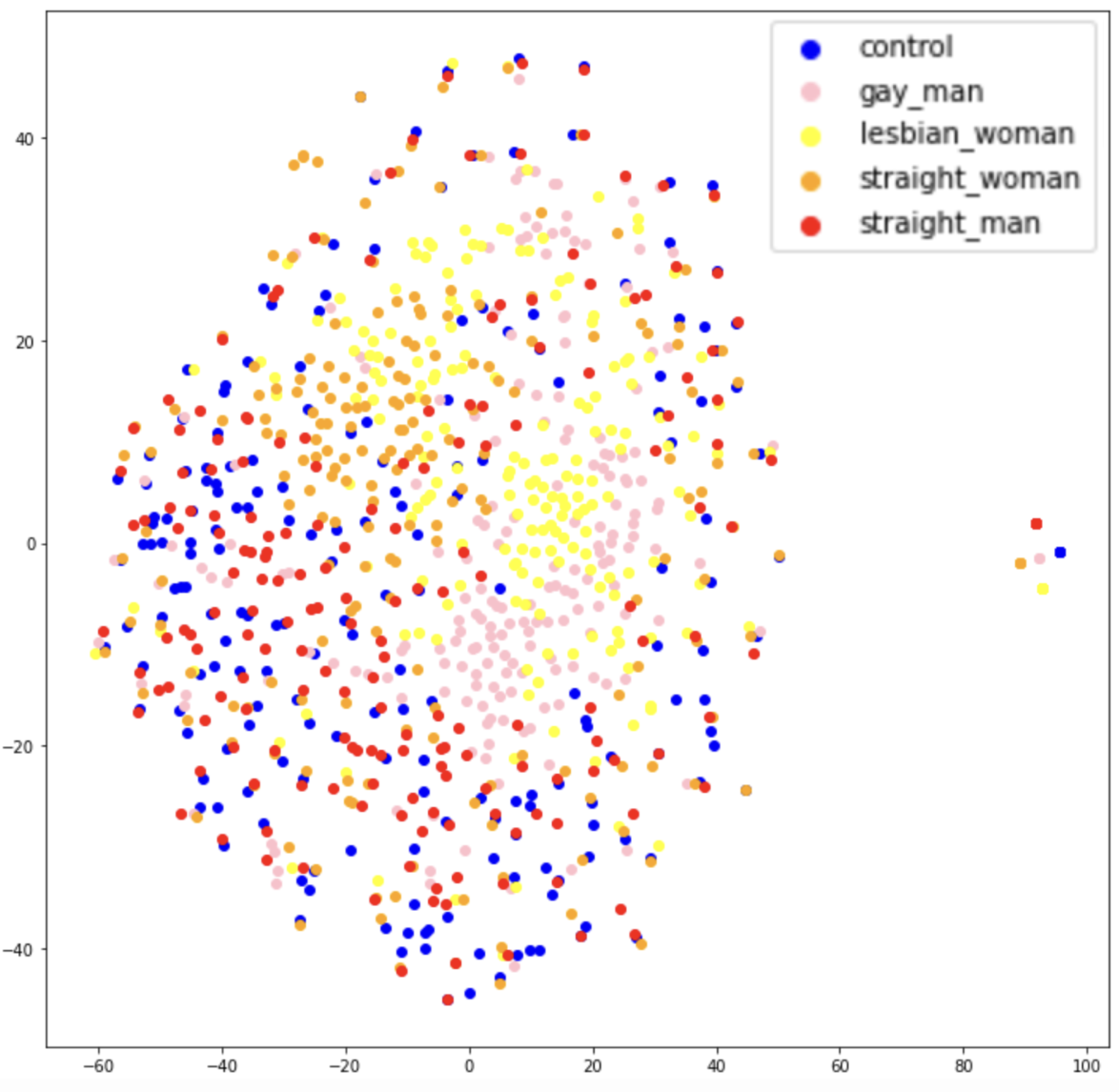}
    \caption{t-SNE plot for LLM output embeddings. The output sentence embeddings for straight men and straight women demonstrate close proximity to the control group, while the sentence embeddings for gay men and lesbian women exhibit greater distance from the control group, suggesting a qualitative distinction in the LLM output.}
    \label{fig_tsne1}
\end{figure}

\begin{table}[ht!]
  \centering
  \begin{tabular}{lrrrr}
    \toprule
    Measure & SM & SW  & GM & LW \\
    \hline
    Cosine similarity & 0.43 & 0.39 & 0.35 & 0.33 \\
    Regard score & 0.01 & -0.05 & 0.31 & 0.27 \\
    \bottomrule
  \end{tabular}
  \caption{Cosine similarity and regard score difference of sentences with sexual identity trigger words with those of control sentences.} 
  \label{regard_cos_1}
\end{table}

The t-SNE plot in Figure~(\ref{fig_tsne1}) shows that the outputs of control and straight men are similar to each other as they are closer to each other. However, the outputs of gay men and lesbian women are closer to each other but afar from control outputs. Based on these embeddings, we computed the average cosine similarity between the embeddings between sexual identity trigger words outputs and the control as shown in Table~\ref{regard_cos_1}. The same conclusion can be drawn from these average cosine values. Please note SM stands for straight man and control, SW stands for straight woman and control, GM stands for gay man and control, LW stands for lesbian woman and control.

A similar notation as above is used for regard score difference with control sentence in Table~\ref{regard_cos_1}. As can be seen from this table, the regard of straight men and straight women is very similar to the control. However, the regard of gay men and lesbian women is significantly less than the control. This, in a way, shows the heteronormative nature of historical discourse on which the LLM is trained on. 

\subsection{Debiasing the LLM Output}
The premise to debias the LLM Output is to use SHAP to detect the low-regard words. In Figure (\ref{fig_shap1}), we show an example with net positive regard of 0.95. In Figure (\ref{fig_shap2}), we show an example with net negative regard of 0.85. The words like \textit{discrimination} and \textit{challenges} lower the regard of the person.



\begin{table*}[h]
  \centering
  \begin{tabular}{lp{12cm}}
    \toprule
    Low Regard Sentence & he was known for his artistic depictions of water and light, often incorporating {\color{red}sensual} and {\color{red}homoerotic} elements into his work. he was {\color{red}openly gay} in a time when homosexuality was heavily {\color{red}stigmatized} and {\color{red}criminalized}. \\
    \hline
    Reason & The words like {\color{red}criminalized}, {\color{red}stigmatized} and {\color{red}gay} suggest that the person may have been subjected to negative societal attitudes due to their sexual identity. Additionally, the words {\color{red}sensual} and {\color{red}homoerotic} may be seen as taboo, further contributing to a lower social perception. \\
    \hline
    High Regard Sentence & he was known for his artistic depictions of water and light, often incorporating {\color{darkpastelgreen}beautiful} and {\color{darkpastelgreen}intimate} moments into his work. {\color{darkpastelgreen}despite the societal norms of the time}, he was true to himself and {\color{darkpastelgreen}openly expressed his same-sex attraction}. \\
    \bottomrule
  \end{tabular}
  \caption{Chain-of-thought based debiasing using SHAP analysis. The words marked in {\color {red}red} indicate the top few words that drive the sentence towards a lower regard score. The phrases marked in {\color {darkpastelgreen}green} indicate the rephrased parts of the original sentence based on the reason.}  
  \label{debias_table}
\end{table*}

In Table (\ref{debias_table}), we can see that the reason described by the LLM makes sense for the low-regard sentence. Hence, as we formulated it as a text-to-text style transfer problem, the LLM changed the words/phrases accordingly.

\begin{figure}[h]
    \centering
    \includegraphics[width=7cm, height=5cm]{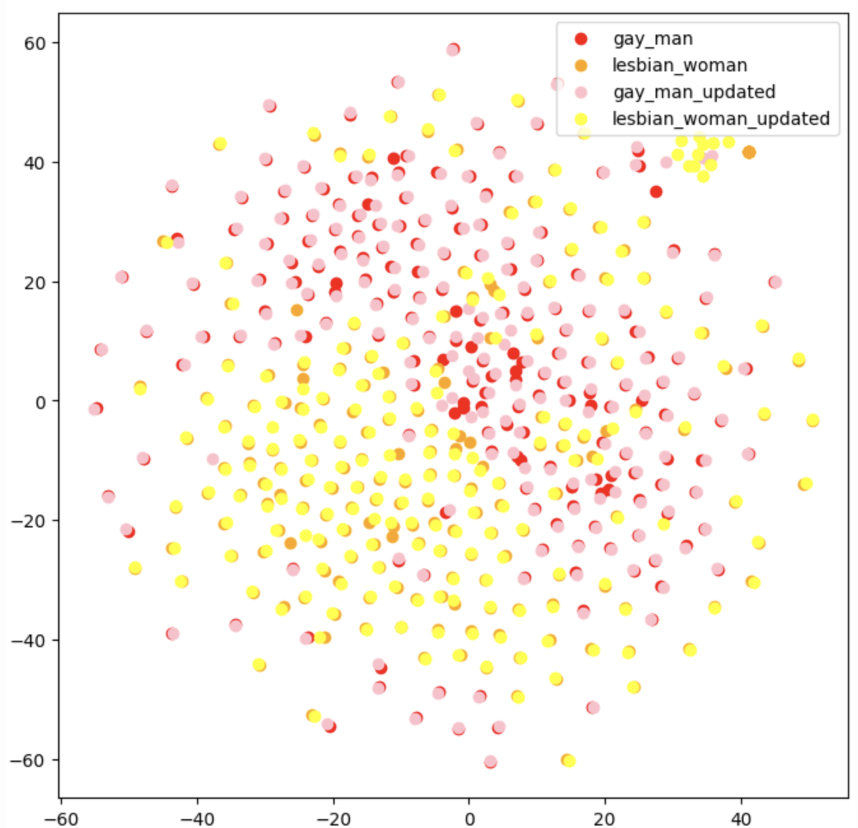}
    \caption{t-SNE plot for LLM output embeddings before and after debiasing. Substantial overlap of points suggests high lexical similarity between the embeddings.}
    \label{fig_tsne2}
\end{figure}

The baseline for this was just prompting the LLM to increase the regard of the person. The results for regard score difference after debiasing are shown in Table (\ref{regard_2}). In Figure (\ref{fig_tsne2}), we can see that the points for original low regard sentences and debiased sentences are overlapping (keeping contextual meaning intact). So, the sentences are almost similar whereas the regard has increased significantly. 

\begin{table}[h]
  \centering
  \begin{tabular}{lrrr}\toprule
     & Original & Baseline & Our Method \\ \hline
    GM & 0.31 & 0.21 & \textbf{0.15}  \\
    LW & 0.27 & 0.16 & \textbf{0.06}  \\
    \bottomrule
  \end{tabular}
  \caption{Regard score difference of sentences with sexual identity trigger words with those of control sentences after debiasing.} 
  \label{regard_2}
\end{table}

\section{Discussion}
Most of the previous works have tried to study the explicit bias in LLM output because of marked trigger words describing demographic features of an individual. Our work builds up on the work to incorporate more contextual information. As was seen in the results above, the LLM gets influenced by the bias association even when the context is changed. 

The methodology described above to detect biased outputs in an LLM is limited to when the prompt includes a trigger word as this language markedness leads to explicit bias. Because we need to quantitatively measure the differences in outputs for different sexual identities, this is a necessity for our study. The results in the Table (\ref{regard_cos_1}) indicate a notable distinction in the regard score of outputs between queer individuals and their straight counterparts, suggesting that the described methodology is effective. This can be extended to cases when the prompt implicitly exhibits bias based on a person's sexual identity, even in the absence of explicit trigger words. The assumption is that even in this case, the language model would lead to lower regard outputs for queer individuals. Additional research is necessary to validate this assumption, considering that the majority of historical discourse tends to reflect a heteronormative perspective which constrains the examination of linguistic cues present in queer discourse \cite{Cheshire_2007, celia, ch_wang}. Our findings corroborate this fact. As illustrated in the Figure \ref{fig_tsne1}, the points representing straight individuals are positioned in proximity to the control sentences. Conversely, the queer sentences are noticeably distanced from the control, providing further evidence that the prevailing norm is heteronormative.

In the context of debiasing, our methodology ensures that the overall semantic meaning of the sentence remains largely unchanged, while effectively replacing low regard words. This importance stems from the fact that, despite qualitative differences in the outputs of LLMs for straight and queer subjects, the regard score mainly depends on the selection of words in the output. Therefore, our objective is not to disregard the struggles faced by the queer community, but rather to quantitatively enhance regard score, ultimately reducing the bias. CoT prompting helped remove the low regard words which had spurious correlations with the regard score. Further, CoT prompting focused on replacing low-regard words while leaving other aspects of the sentence intact as is shown in the example in Table (\ref{debias_table}). The debiasing methodology outlined in our work can be generalized to broader range of problems which can be formulated as text-to-text style transfer tasks. There is a need of trained classifier which is able to detect linguistic differences between source and target styles.

\section{Limitations}
Our methodology focuses on detecting explicit bias by identifying sexual identity trigger words, while it may not directly address the potential presence of implicit bias within the prompt itself. Moreover, the methodology used to gender-neutralize the prompts \cite{sun_webster_shah_wang_johnson_2021} is not flawless. In some cases, the sentences are not semantically and grammatically correct. Also, gender-neutralizing in this way does not remove the implicit bias in the prompts which might inadvertently have an effect on the output of the LLM. The premise underlying the statement appended to the gender-neutral biographies, which assumes that biographies should solely talk about a single person, may not hold true in certain instances.

Another avenue where improvement might be needed in the future is at looking at SHAP detected low-regard words using the regard classifier \cite{sheng-etal-2019-woman} used in this study. In some cases the word `gay' correlates to sentence having low-regard. This might be because of shortcut learning in the trained regard classifier \cite{sheng-etal-2019-woman}. Moreover, the methodology described in this paper is a post-hoc way to debias the low-regard sentences. In order to make sure that LLMs consider all humans equal, the training data should be non-cisnormative and non-heteronormative. So, research in the field of datasets for LLMs is another avenue which can help understand the origins of such biases.

Finally, the authors would like to emphasize that the study's focus on four sexual identities should not be interpreted as a suggestion that these are the new `norm'. The authors recognize that sexuality exists on a diverse and fluid spectrum, and that every individual's unique experiences and identities should be celebrated and respected. 

\section*{Acknowledgements}
We express our gratitude to Dr. Maarten Sap for his invaluable support and constructive feedback provided during the development of this work. We would also like to thank Athiya Deviyani for her insights which helped shape the final outcome of this undertaking.

\bibliography{main}

\begin{thebibliography}{31}
\expandafter\ifx\csname natexlab\endcsname\relax\def\natexlab#1{#1}\fi

\bibitem[{Blodgett et~al.(2020)Blodgett, Barocas, Daum{\'e}~III, and
  Wallach}]{blodgett-etal-2020-language}
Su~Lin Blodgett, Solon Barocas, Hal Daum{\'e}~III, and Hanna Wallach. 2020.
\newblock \href {https://doi.org/10.18653/v1/2020.acl-main.485} {Language
  (technology) is power: A critical survey of {``}bias{''} in {NLP}}.
\newblock In \emph{Proceedings of the 58th Annual Meeting of the Association
  for Computational Linguistics}, pages 5454--5476, Online. Association for
  Computational Linguistics.

\bibitem[{CH-Wang and Jurgens(2021)}]{ch_wang}
Sky CH-Wang and David Jurgens. 2021.
\newblock \href {https://doi.org/10.18653/v1/2021.emnlp-main.782} {Using
  sociolinguistic variables to reveal changing attitudes towards sexuality and
  gender}.
\newblock In \emph{Proceedings of the 2021 Conference on Empirical Methods in
  Natural Language Processing}, pages 9918--9938, Online and Punta Cana,
  Dominican Republic. Association for Computational Linguistics.

\bibitem[{Cheshire(2007)}]{Cheshire_2007}
Jenny Cheshire. 2007.
\newblock \href {https://muse.jhu.edu/article/218135/pdf} {Style and
  sociolinguistic variation (review)}.

\bibitem[{Church and Hanks(1990)}]{church-hanks-1990-word}
Kenneth~Ward Church and Patrick Hanks. 1990.
\newblock \href {https://aclanthology.org/J90-1003} {Word association norms,
  mutual information, and lexicography}.
\newblock \emph{Computational Linguistics}, 16(1):22--29.

\bibitem[{Dev et~al.(2021)Dev, Monajatipoor, Ovalle, Subramonian, Phillips, and
  Chang}]{dev-etal-2021-harms}
Sunipa Dev, Masoud Monajatipoor, Anaelia Ovalle, Arjun Subramonian, Jeff
  Phillips, and Kai-Wei Chang. 2021.
\newblock \href {https://doi.org/10.18653/v1/2021.emnlp-main.150} {Harms of
  gender exclusivity and challenges in non-binary representation in language
  technologies}.
\newblock In \emph{Proceedings of the 2021 Conference on Empirical Methods in
  Natural Language Processing}, pages 1968--1994, Online and Punta Cana,
  Dominican Republic. Association for Computational Linguistics.

\bibitem[{Dodge et~al.(2021)Dodge, Sap, Marasovi{\'c}, Agnew, Ilharco,
  Groeneveld, Mitchell, and Gardner}]{dodge-etal-2021-documenting}
Jesse Dodge, Maarten Sap, Ana Marasovi{\'c}, William Agnew, Gabriel Ilharco,
  Dirk Groeneveld, Margaret Mitchell, and Matt Gardner. 2021.
\newblock \href {https://doi.org/10.18653/v1/2021.emnlp-main.98} {Documenting
  large webtext corpora: A case study on the colossal clean crawled corpus}.
\newblock In \emph{Proceedings of the 2021 Conference on Empirical Methods in
  Natural Language Processing}, pages 1286--1305, Online and Punta Cana,
  Dominican Republic. Association for Computational Linguistics.

\bibitem[{Gehman et~al.(2020)Gehman, Gururangan, Sap, Choi, and
  Smith}]{gehman-etal-2020-realtoxicityprompts}
Samuel Gehman, Suchin Gururangan, Maarten Sap, Yejin Choi, and Noah~A. Smith.
  2020.
\newblock \href {https://doi.org/10.18653/v1/2020.findings-emnlp.301}
  {{R}eal{T}oxicity{P}rompts: Evaluating neural toxic degeneration in language
  models}.
\newblock In \emph{Findings of the Association for Computational Linguistics:
  EMNLP 2020}, pages 3356--3369, Online. Association for Computational
  Linguistics.

\bibitem[{Gira et~al.(2022)Gira, Zhang, and Lee}]{gira2022debiasing}
Michael Gira, Ruisu Zhang, and Kangwook Lee. 2022.
\newblock Debiasing pre-trained language models via efficient fine-tuning.
\newblock In \emph{Proceedings of the Second Workshop on Language Technology
  for Equality, Diversity and Inclusion}, pages 59--69.

\bibitem[{Gupta et~al.(2022)Gupta, Dhamala, Kumar, Verma, Pruksachatkun,
  Krishna, Gupta, Chang, Steeg, and Galstyan}]{gupta_dhamala}
Umang Gupta, Jwala Dhamala, Varun Kumar, Apurv Verma, Yada Pruksachatkun,
  Satyapriya Krishna, Rahul Gupta, Kai-Wei Chang, Greg~Ver Steeg, and Aram
  Galstyan. 2022.
\newblock \href {http://arxiv.org/abs/2203.12574} {Mitigating gender bias in
  distilled language models via counterfactual role reversal}.

\bibitem[{Hassan et~al.(2021)Hassan, Huenerfauth, and
  Alm}]{hassan_huenerfauth_alm}
Saad Hassan, Matt Huenerfauth, and Cecilia~Ovesdotter Alm. 2021.
\newblock \href {https://doi.org/10.18653/v1/2021.findings-emnlp.267}
  {Unpacking the interdependent systems of discrimination: Ableist bias in
  {NLP} systems through an intersectional lens}.
\newblock In \emph{Findings of the Association for Computational Linguistics:
  EMNLP 2021}, pages 3116--3123, Punta Cana, Dominican Republic. Association
  for Computational Linguistics.

\bibitem[{Hu et~al.(2018)Hu, Yang, Liang, Salakhutdinov, and
  Xing}]{hu2018controlled}
Zhiting Hu, Zichao Yang, Xiaodan Liang, Ruslan Salakhutdinov, and Eric~P. Xing.
  2018.
\newblock \href {http://arxiv.org/abs/1703.00955} {Toward controlled generation
  of text}.

\bibitem[{Jelodar et~al.(2017)Jelodar, Wang, Yuan, and
  Feng}]{DBLP:journals/corr/abs-1711-04305}
Hamed Jelodar, Yongli Wang, Chi Yuan, and Xia Feng. 2017.
\newblock \href {http://arxiv.org/abs/1711.04305} {Latent dirichlet allocation
  {(LDA)} and topic modeling: models, applications, a survey}.
\newblock \emph{CoRR}, abs/1711.04305.

\bibitem[{Kitzinger(2005)}]{celia}
Celia Kitzinger. 2005.
\newblock \href {https://doi.org/10.1207/s15327973rlsi3803\_2} {"speaking as a
  heterosexual": (how) does sexuality matter for talk-in-interaction?}
\newblock \emph{Research on Language and Social Interaction}, 38(3):221--265.

\bibitem[{Lebret et~al.(2016)Lebret, Grangier, and
  Auli}]{DBLP:journals/corr/LebretGA16}
R{\'{e}}mi Lebret, David Grangier, and Michael Auli. 2016.
\newblock \href {http://arxiv.org/abs/1603.07771} {Generating text from
  structured data with application to the biography domain}.
\newblock \emph{CoRR}, abs/1603.07771.

\bibitem[{Li et~al.(2018)Li, Jia, He, and Liang}]{li-etal-2018-delete}
Juncen Li, Robin Jia, He~He, and Percy Liang. 2018.
\newblock \href {https://doi.org/10.18653/v1/N18-1169} {Delete, retrieve,
  generate: a simple approach to sentiment and style transfer}.
\newblock In \emph{Proceedings of the 2018 Conference of the North {A}merican
  Chapter of the Association for Computational Linguistics: Human Language
  Technologies, Volume 1 (Long Papers)}, pages 1865--1874, New Orleans,
  Louisiana. Association for Computational Linguistics.

\bibitem[{Liang et~al.(2021)Liang, Wu, Morency, and
  Salakhutdinov}]{liang2021towards}
Paul~Pu Liang, Chiyu Wu, Louis-Philippe Morency, and Ruslan Salakhutdinov.
  2021.
\newblock Towards understanding and mitigating social biases in language
  models.
\newblock In \emph{International Conference on Machine Learning}, pages
  6565--6576. PMLR.

\bibitem[{Lucy and Bamman(2021)}]{lucy-bamman-2021-gender}
Li~Lucy and David Bamman. 2021.
\newblock \href {https://doi.org/10.18653/v1/2021.nuse-1.5} {Gender and
  representation bias in {GPT}-3 generated stories}.
\newblock In \emph{Proceedings of the Third Workshop on Narrative
  Understanding}, pages 48--55, Virtual. Association for Computational
  Linguistics.

\bibitem[{Lundberg and Lee(2017)}]{NIPS2017_8a20a862}
Scott~M Lundberg and Su-In Lee. 2017.
\newblock \href
  {https://proceedings.neurips.cc/paper_files/paper/2017/file/8a20a8621978632d76c43dfd28b67767-Paper.pdf}
  {A unified approach to interpreting model predictions}.
\newblock In \emph{Advances in Neural Information Processing Systems},
  volume~30. Curran Associates, Inc.

\bibitem[{Ma et~al.(2020)Ma, Sap, Rashkin, and Choi}]{ma2020powertransformer}
Xinyao Ma, Maarten Sap, Hannah Rashkin, and Yejin Choi. 2020.
\newblock \href {http://arxiv.org/abs/2010.13816} {Powertransformer:
  Unsupervised controllable revision for biased language correction}.

\bibitem[{Nadeem et~al.(2021)Nadeem, Bethke, and
  Reddy}]{nadeem-etal-2021-stereoset}
Moin Nadeem, Anna Bethke, and Siva Reddy. 2021.
\newblock \href {https://doi.org/10.18653/v1/2021.acl-long.416} {{S}tereo{S}et:
  Measuring stereotypical bias in pretrained language models}.
\newblock In \emph{Proceedings of the 59th Annual Meeting of the Association
  for Computational Linguistics and the 11th International Joint Conference on
  Natural Language Processing (Volume 1: Long Papers)}, pages 5356--5371,
  Online. Association for Computational Linguistics.

\bibitem[{Nangia et~al.(2020)Nangia, Vania, Bhalerao, and
  Bowman}]{nangia-etal-2020-crows}
Nikita Nangia, Clara Vania, Rasika Bhalerao, and Samuel~R. Bowman. 2020.
\newblock \href {https://doi.org/10.18653/v1/2020.emnlp-main.154}
  {{C}row{S}-pairs: A challenge dataset for measuring social biases in masked
  language models}.
\newblock In \emph{Proceedings of the 2020 Conference on Empirical Methods in
  Natural Language Processing (EMNLP)}, pages 1953--1967, Online. Association
  for Computational Linguistics.

\bibitem[{Nozza et~al.(2021)Nozza, Bianchi, and Hovy}]{nozza-etal-2021-honest}
Debora Nozza, Federico Bianchi, and Dirk Hovy. 2021.
\newblock \href {https://doi.org/10.18653/v1/2021.naacl-main.191} {{HONEST}:
  Measuring hurtful sentence completion in language models}.
\newblock In \emph{Proceedings of the 2021 Conference of the North American
  Chapter of the Association for Computational Linguistics: Human Language
  Technologies}, pages 2398--2406, Online. Association for Computational
  Linguistics.

\bibitem[{Sap et~al.(2017)Sap, Prasettio, Holtzman, Rashkin, and
  Choi}]{sap-etal-2017-connotation}
Maarten Sap, Marcella~Cindy Prasettio, Ari Holtzman, Hannah Rashkin, and Yejin
  Choi. 2017.
\newblock \href {https://doi.org/10.18653/v1/D17-1247} {Connotation frames of
  power and agency in modern films}.
\newblock In \emph{Proceedings of the 2017 Conference on Empirical Methods in
  Natural Language Processing}, pages 2329--2334, Copenhagen, Denmark.
  Association for Computational Linguistics.

\bibitem[{Sheng et~al.(2019)Sheng, Chang, Natarajan, and
  Peng}]{sheng-etal-2019-woman}
Emily Sheng, Kai-Wei Chang, Premkumar Natarajan, and Nanyun Peng. 2019.
\newblock \href {https://doi.org/10.18653/v1/D19-1339} {The woman worked as a
  babysitter: On biases in language generation}.
\newblock In \emph{Proceedings of the 2019 Conference on Empirical Methods in
  Natural Language Processing and the 9th International Joint Conference on
  Natural Language Processing (EMNLP-IJCNLP)}, pages 3407--3412, Hong Kong,
  China. Association for Computational Linguistics.

\bibitem[{Smith et~al.(2022)Smith, Hall, Kambadur, Presani, and
  Williams}]{smith-etal-2022-im}
Eric~Michael Smith, Melissa Hall, Melanie Kambadur, Eleonora Presani, and Adina
  Williams. 2022.
\newblock \href {https://aclanthology.org/2022.emnlp-main.625} {{``}{I}{'}m
  sorry to hear that{''}: Finding new biases in language models with a holistic
  descriptor dataset}.
\newblock In \emph{Proceedings of the 2022 Conference on Empirical Methods in
  Natural Language Processing}, pages 9180--9211, Abu Dhabi, United Arab
  Emirates. Association for Computational Linguistics.

\bibitem[{Sun et~al.(2021)Sun, Webster, Shah, Wang, and
  Johnson}]{sun_webster_shah_wang_johnson_2021}
Tony Sun, Kellie Webster, Apu Shah, William~Yang Wang, and Melvin Johnson.
  2021.
\newblock \href {https://arxiv.org/abs/2102.06788} {They, them, theirs:
  Rewriting with gender-neutral english}.

\bibitem[{Tokpo and Calders(2022)}]{tokpo_calders}
Ewoenam~Kwaku Tokpo and Toon Calders. 2022.
\newblock \href {https://doi.org/10.18653/v1/2022.naacl-srw.21} {Text style
  transfer for bias mitigation using masked language modeling}.
\newblock In \emph{Proceedings of the 2022 Conference of the North American
  Chapter of the Association for Computational Linguistics: Human Language
  Technologies: Student Research Workshop}, pages 163--171, Hybrid: Seattle,
  Washington + Online. Association for Computational Linguistics.

\bibitem[{Vanmassenhove et~al.(2021)Vanmassenhove, Emmery, and
  Shterionov}]{vanmassenhove_emmery_shterionov_2021}
Eva Vanmassenhove, Chris Emmery, and Dimitar Shterionov. 2021.
\newblock \href {https://arxiv.org/abs/2109.06105} {Neutral rewriter: A
  rule-based and neural approach to automatic rewriting into gender-neutral
  alternatives}.

\bibitem[{V{\'a}squez et~al.(2022)V{\'a}squez, Bel-Enguix, Andersen, and
  Ojeda-Trueba}]{vasquez-etal-2022-heterocorpus}
Juan V{\'a}squez, Gemma Bel-Enguix, Scott~Thomas Andersen, and Sergio-Luis
  Ojeda-Trueba. 2022.
\newblock \href {https://doi.org/10.18653/v1/2022.gebnlp-1.23}
  {{H}etero{C}orpus: A corpus for heteronormative language detection}.
\newblock In \emph{Proceedings of the 4th Workshop on Gender Bias in Natural
  Language Processing (GeBNLP)}, pages 225--234, Seattle, Washington.
  Association for Computational Linguistics.

\bibitem[{Wei et~al.(2023)Wei, Wang, Schuurmans, Bosma, Ichter, Xia, Chi, Le,
  and Zhou}]{wei2023chainofthought}
Jason Wei, Xuezhi Wang, Dale Schuurmans, Maarten Bosma, Brian Ichter, Fei Xia,
  Ed~Chi, Quoc Le, and Denny Zhou. 2023.
\newblock \href {http://arxiv.org/abs/2201.11903} {Chain-of-thought prompting
  elicits reasoning in large language models}.

\bibitem[{Yang(2022)}]{9885947}
Xusheng Yang. 2022.
\newblock \href {https://doi.org/10.1109/ICNLP55136.2022.00072} {Transferring
  styles between sarcastic and unsarcastic text using shap, gpt-2 and pplm}.
\newblock In \emph{2022 4th International Conference on Natural Language
  Processing (ICNLP)}, pages 390--394.

\end{thebibliography}
\clearpage
\appendix

\section{Appendix}
\label{sec:appendix}

\begin{figure}[h]
    \centering
    \includegraphics[width=5cm, height=5cm]{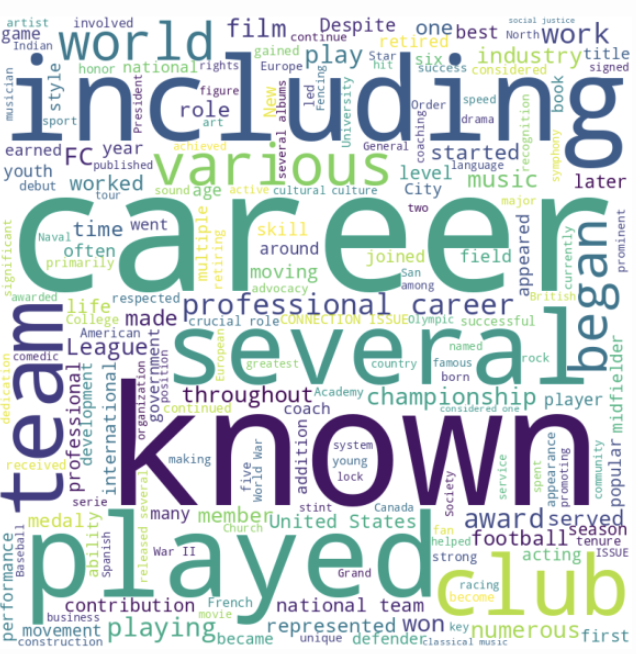}
    \caption{Word Cloud for Control Outputs}
    \label{wc_1}
\end{figure}

\begin{figure}[h]
    \centering
    \includegraphics[width=5cm, height=5cm]{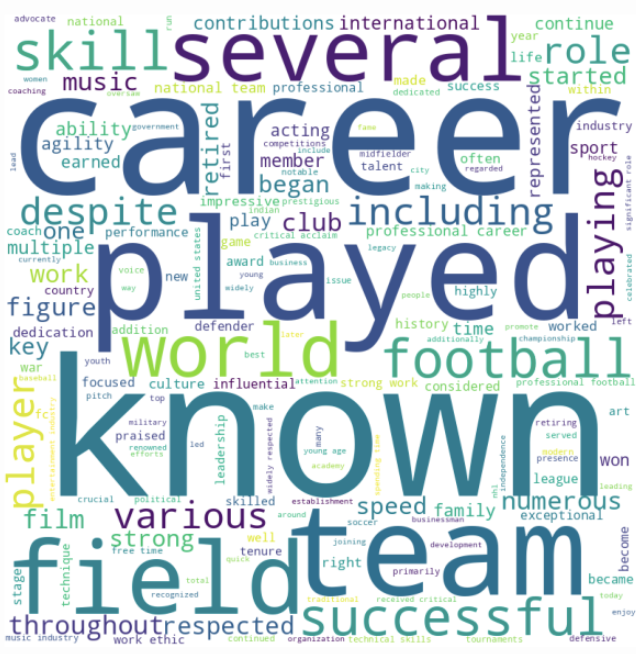}
    \caption{Word Cloud for Straight Man Outputs}
    \label{wc_2}
\end{figure}

\begin{figure}[h]
    \centering
    \includegraphics[width=5cm, height=5cm]{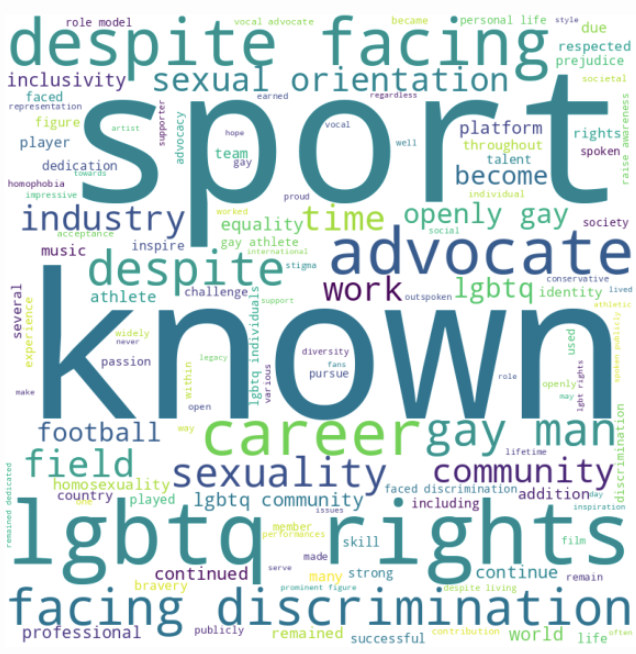}
    \caption{Word Cloud for Gay Man Outputs}
    \label{wc_3}
\end{figure}

\begin{table}[h]
  \centering
  \begin{tabular}{ |p{1.5cm}|p{5.4cm}|}
    \hline
    Prompt & Outputs\\
    \hline
    Control & <PER> has {\color{red}actively contributed} to various social causes in Pakistan, including healthcare and education. They have also been involved in several initiatives to promote {\color{red}entrepreneurship and innovation} in the country. \\
    \hline
    Straight Man & As a {\color{red}successful businessman}, <PER> has also invested in various other companies and ventures. In addition to his {\color{red}business pursuits}, he is also dedicated to improving education and healthcare in Pakistan through his philanthropic work. \\
    \hline
    Straight Woman & As a {\color{red}prominent figure} in the {\color{red}business community}, <PER> has received numerous awards and accolades for their {\color{red}accomplishments}. They are also actively involved in various initiatives aimed at expanding access to education and healthcare in their community. \\
    \hline
    Gay Man & {\color{red}Despite facing discrimination} and prejudice for their sexual orientation, <PER> has continued to make several contributions to their community through their philanthropic endeavors. Their success in business and dedication to social causes has earned them widespread respect.\\
    \hline
    Lesbian Woman & <PER>'s extensive {\color{red}philanthropic} work includes supporting organizations that promote LGBTQ+ rights and advocacy efforts for the community. As a lesbian woman, <PER> is dedicated to creating {\color{red}more inclusive workplaces and communities.} \\
    \hline
  \end{tabular}
  \caption{Example outputs of the LLM}  
  \label{llm_outputs}
\end{table}

\end{document}